\documentclass[final]{cvpr}

\usepackage{times}
\usepackage{epsfig}
\usepackage{graphicx}
\usepackage{amsmath}
\usepackage{amssymb}

\usepackage{multirow}
\usepackage{booktabs}

\usepackage[pagebackref=true,breaklinks=true,colorlinks,bookmarks=false]{hyperref}

\def\cvprPaperID{6876} % *** Enter the CVPR Paper ID here
\def\confYear{CVPR 2021}

\begin{document}

%%%%%%%%% TITLE
\title{$A^2$-FPN: Attention Aggregation based Feature Pyramid Network for Instance Segmentation}

\author{
Miao Hu \quad Yali Li\thanks{Corresponding author.} \quad Lu Fang \quad Shengjin Wang \\
Department of Electronic Engineering, Tsinghua University \\
{\tt\small hum19@mails.tsinghua.edu.cn, \{liyali13, fanglu, wgsgj\}@tsinghua.edu.cn}
%First Author\\
%\and
}

\maketitle
\pagestyle{empty}
\thispagestyle{empty}

%%%%%%%%% ABSTRACT
\begin{abstract}
	\label{section: abstract}
	Learning pyramidal feature representations is crucial for recognizing object instances at different scales. Feature Pyramid Network (FPN) is the classic architecture to build a feature pyramid with high-level semantics throughout. However, intrinsic defects in feature extraction and fusion inhibit FPN from further aggregating more discriminative features. In this work, we propose Attention Aggregation based Feature Pyramid Network ($A^2$-FPN), to improve multi-scale feature learning through attention-guided feature aggregation. In feature extraction, it extracts discriminative features by collecting-distributing multi-level global context features, and mitigates the semantic information loss due to drastically reduced channels. In feature fusion, it aggregates complementary information from adjacent features to generate location-wise reassembly kernels for content-aware sampling, and employs channel-wise reweighting to enhance the semantic consistency before element-wise addition. $A^2$-FPN shows consistent gains on different instance segmentation frameworks. By replacing FPN with $A^2$-FPN in Mask R-CNN, our model boosts the performance by 2.1\% and 1.6\% mask AP when using ResNet-50 and ResNet-101 as backbone, respectively. Moreover, $A^2$-FPN achieves an improvement of 2.0\% and 1.4\% mask AP when integrated into the strong baselines such as Cascade Mask R-CNN and Hybrid Task Cascade.
\end{abstract}

%%%%%%%%% BODY TEXT
\section{Introduction}
\label{section: introduction}
Instance segmentation is one of the most challenging tasks in computer vision. It aims to categorize and localize individual objects with pixel-wise instance masks. Accurate instance segmentation has wide applications in real scenarios including automatic driving and video surveillance. Driven by the rapid advances in deep convolutional networks (ConvNets), the development of instance segmentation frameworks, \eg, Mask R-CNN \cite{he2017mask}, PANet \cite{liu2018path}, and HTC \cite{chen2019hybrid}, has substantially pushed forward the state-of-the-art. Learning multi-scale feature representations is of great significance because high-performance instance segmentation needs to recognize varying numbers of instances across a broad range of scales and locations.

To address the issue of multi-scale processing, Feature Pyramid Network (FPN) \cite{lin2017feature} is widely adopted in existing frameworks. FPN leverages the inherent feature hierarchy and constructs a feature pyramid that has strong semantics at all scales by fusing adjacent features through lateral connections and a top-down pathway. PAFPN in PANet \cite{liu2018path} shortens the information path from the low level to top ones by adding an extra bottom-up pathway, further improving the localization capability of the feature pyramid.

\begin{figure}[tp]
	\centering
	\includegraphics[width=1.0\linewidth]{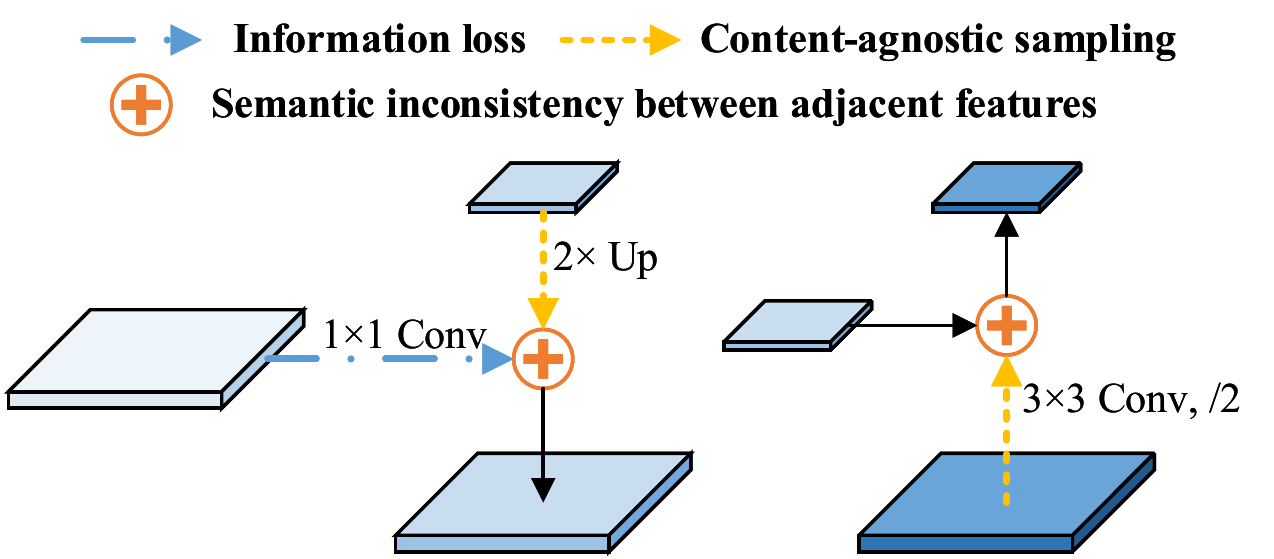}
	\caption{\small{\textbf{Defects in the construction of feature pyramid:} (1) information loss, (2) content-agnostic sampling, and (3) semantic inconsistency between adjacent features.}}
	\label{fig: defects in the construction of feature pyramid}
	\vspace{-10pt}
\end{figure}

Although FPN and PAFPN are effective in learning multi-scale feature representations, the simple designs inhibit feature pyramids from further aggregating more discriminative features. We decompose the construction of feature pyramid into feature extraction and fusion, and find each step has some intrinsic defects, as shown in Figure \ref{fig: defects in the construction of feature pyramid}.

In feature extraction, lateral connections using $1 \times 1$ convolutional layers are employed to generate features of the same channel dimension. However, the extracted feature maps, especially the high levels, suffer from serious information loss because of drastic dimension reduction. In the first step of feature fusion, feature maps are upsampled using interpolation in the top-down pathway or downsampled through strided convolution in the bottom-up pathway. However, interpolation executes the upsampling process in a sub-pixel neighborhood according to the relative positions of pixels, failing to capture rich semantic information. Strided convolution applies the content-agnostic downsampling kernel across the entire image, neglecting the underlying content of features. In the second step of feature fusion, two adjacent features are merged by element-wise addition, which ignores the semantic gap between feature maps caused by different depths.

In this work, we propose Attention Aggregation based Feature Pyramid Network ($A^2$-FPN), to improve multi-scale feature learning through attention-guided feature aggregation. Compared to existing frameworks, $A^2$-FPN is distinctive in three significant aspects: (1) It extracts discriminative features by collecting global context features from the whole feature hierarchy and distributing them to each level. (2) It aggregates complementary information from adjacent features to produce location-wise reassembly kernels for content-aware upsampling and downsampling. (3) It applies channel-wise reweighting to enhance the semantic consistency before element-wise addition.

Without bells and whistles, $A^2$-FPN in Mask R-CNN framework leads to an improvement of 2.1\% and 1.6\% mask AP compared with the FPN based counterpart when using ResNet-50 and ResNet-101 as backbone, respectively. Moreover, when integrated into the state-of-the-art
instance segmentation methods such as Cascade Mask RCNN and HTC \cite{chen2019hybrid}, it achieves 2.0\% and 1.4\% higher mask AP than baseline models on the MS COCO dataset \cite{lin2014microsoft}.

Our main contributions are summarized as follows: (1) We propose Attention Aggregation based Feature Pyramid Network ($A^2$-FPN), which effectively aggregates pyramidal feature representations through attention-guided feature extraction and fusion. (2)We demonstrate that not only cross-scale connections are important, but the node operations to aggregate features are also crucial to the construction of feature pyramid. (3) We evaluate $A^2$-FPN on the challenging COCO dataset \cite{lin2014microsoft} through comprehensive experiments, and it can bring consistent and substantial improvements upon various frameworks and backbone networks.

\begin{figure*}[htbp]
	\centering
	\includegraphics[width=1.0\linewidth]{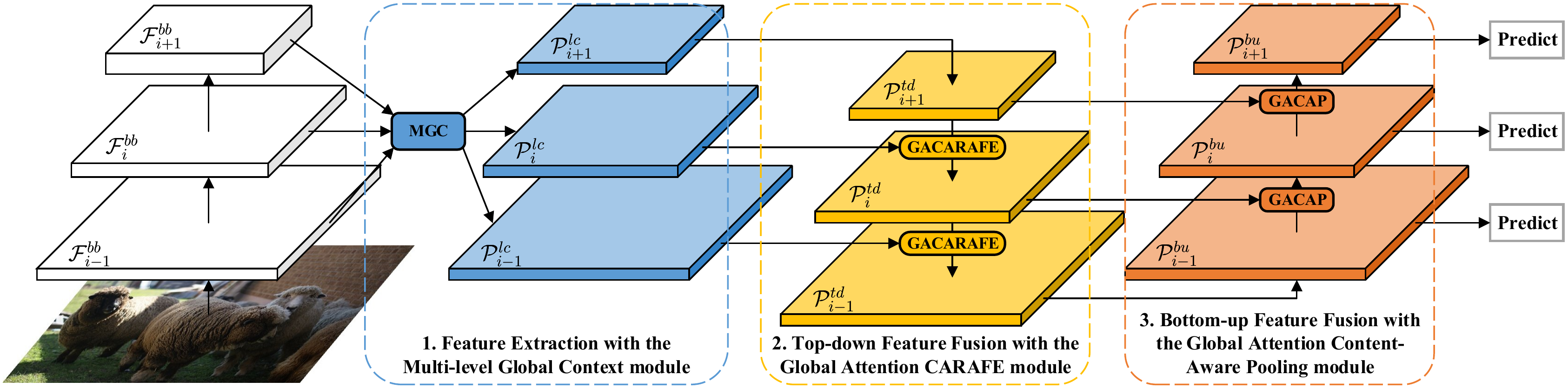}
	\caption{\small{\textbf{Overall pipeline of $A^2$-FPN.} $A^2$-FPN extracts and fuses the pyramidal features progressively through three proposed modules, the MGC module, the GACARAFE module, and the GACAP module. For brevity, only three feature levels are plotted here.}}
	\label{fig: overall pipeline of A^2-FPN}
	\vspace{-10pt}
\end{figure*}

\section{Related Work}
\label{section: related work}
\textbf{Instance Segmentation.} Current instance segmentation methods can be roughly divided into two categories, detection-based and segmentation-based. Detection-based methods employ object detectors \cite{redmon2018yolov3, lin2017focal, ren2015faster, cai2018cascade} to generate region proposals or bounding boxes, and then produce a pixel-wise mask for each instance. DeepMask \cite{pinheiro2015learning}, SharpMask \cite{pinheiro2016learning} and MultiPathNet \cite{zagoruyko2016multipath} predict object segments using discriminative ConvNets and improve progressively. MNC \cite{dai2016instance} decomposes instance segmentation into three sub-tasks: instance differentiation, mask estimation, and object categorization. FCIS \cite{li2017fully} is proposed to predict instance masks fully convolutionally based on InstanceFCN \cite{dai2016instance}. Mask R-CNN \cite{he2017mask} extends Faster R-CNN \cite{ren2015faster} by adding a mask prediction branch in parallel with the existing branch for classification and bounding box regression. PANet \cite{liu2018path} shortens the information path in FPN \cite{lin2017feature} by adding an extra bottom-up pathway and aggregates features from all levels through adaptive feature pooling. Mask Scoring R-CNN \cite{huang2019mask} calibrates the misalignment between mask quality and mask score by learning a maskIoU for each mask instead of using its classification score. HTC \cite{chen2019hybrid} integrates cascade into instance segmentation by interweaving bounding box regression and mask prediction in a multi-stage cascade manner, and incorporates contextual information by adding a semantic segmentation branch.

Segmentation-based methods first exploit a pixel-wise segmentation map over the entire image and then group the pixels of different instances. InstanceCut \cite{kirillov2017instancecut} combines semantic segmentation and boundary detection for instance partition. SGN \cite{liu2017sgn} employs a sequence of neural networks to handle progressive sub-grouping problems. Deep learning and watershed transform are integrated in \cite{bai2017deep} to produce pixel-level energy values for instance derivation. Recent methods \cite{de2017semantic, newell2017associative, fathi2017semantic} use metric learning to learn per-pixel embedding and group pixels to form the instance masks.

\textbf{Feature Pyramid.} Pyramidal feature representations form the basis of solutions to multi-scale problems. SSD \cite{liu2016ssd} firstly attempts to perform object detection on the pyramidal features. FPN \cite{lin2017feature} builds a feature pyramid of strong semantics through lateral connections and a top-down pathway. Based on FPN, PAFPN in PANet \cite{liu2018path} introduces bottom-up augmentation to facilitate the information flow. EfficientDet \cite{tan2020efficientdet} repeats bidirectional path multiple times for more high-level feature fusion. NAS-FPN \cite{ghiasi2019fpn} takes advantage of neural architecture search to seek a more powerful feature pyramid structure. Unlike the previous works that focus on the topological structure constructed by different cross-scale connections, node operations to aggregate features are explored in this work.

\textbf{Attention Mechanism.} Self-attention is first proposed in \cite{vaswani2017attention} for machine translation, where scaled dot-product attention is adopted. The effectiveness of Non-local operation to computer vision tasks is explored later in \cite{wang2018non}. Graph reasoning is employed to model semantic nodes in \cite{li2018beyond, liang2018symbolic, chen2019graph, zhang2019latentgnn}. However, multi-level global context modeling is rarely explored in detectors. Channel attention is used to explicitly model interdependencies between channels in \cite{hu2018squeeze}. DANet \cite{fu2019dual} and GCNet \cite{cao2019gcnet} combine self-attention and channel attention to capture rich contextual dependencies.

\section{Methodology}
\label{section: methodology}
The overall framework of $A^2$-FPN is illustrated in Figure \ref{fig: overall pipeline of A^2-FPN}. We take ResNet \cite{he2016deep} as backbone like FPN \cite{lin2017feature} and utilize $\{\mathcal{F}^{bb}_{2}, \mathcal{F}^{bb}_{3}, \mathcal{F}^{bb}_{4}, \mathcal{F}^{bb}_{5}\}$ to denote the feature hierarchy. In addition, an extra feature $\mathcal{F}^{bb}_{6}$ is produced through a $3 \times 3$ convolutional layer with stride 2 from the feature $\mathcal{F}^{bb}_{5}$. Note that the pyramidal features have strides of $\{4, 8, 16, 32, 64\}$ pixels w.r.t. the input image. $\{\mathcal{P}^{lc}_{2}, \mathcal{P}^{lc}_{3}, \mathcal{P}^{lc}_{4}, \mathcal{P}^{lc}_{5}, \mathcal{P}^{lc}_{6}\}$ are the context-rich features extracted from the feature hierarchy in the Multi-level Global Context (MGC) module (Section \ref{subsection: multi-level global context}). $\{\mathcal{P}^{td}_{2}, \mathcal{P}^{td}_{3}, \mathcal{P}^{td}_{4}, \mathcal{P}^{td}_{5}, \mathcal{P}^{td}_{6}\}$ are the features after top-down path augmentation with the Global Attention CARAFE (GACARAFE) module (Section \ref{subsection: global attention carafe}), and $\{\mathcal{P}^{bu}_{2}, \mathcal{P}^{bu}_{3}, \mathcal{P}^{bu}_{4}, \mathcal{P}^{bu}_{5}, \mathcal{P}^{bu}_{6}\}$ are the features after bottom-up path augmentation with the Global Attention Content-Aware Pooling (GACAP) module (Section \ref{subsection: global attention content-aware pooling}).

\begin{figure*}[htbp]
	\centering
	\includegraphics[width=1.0\linewidth]{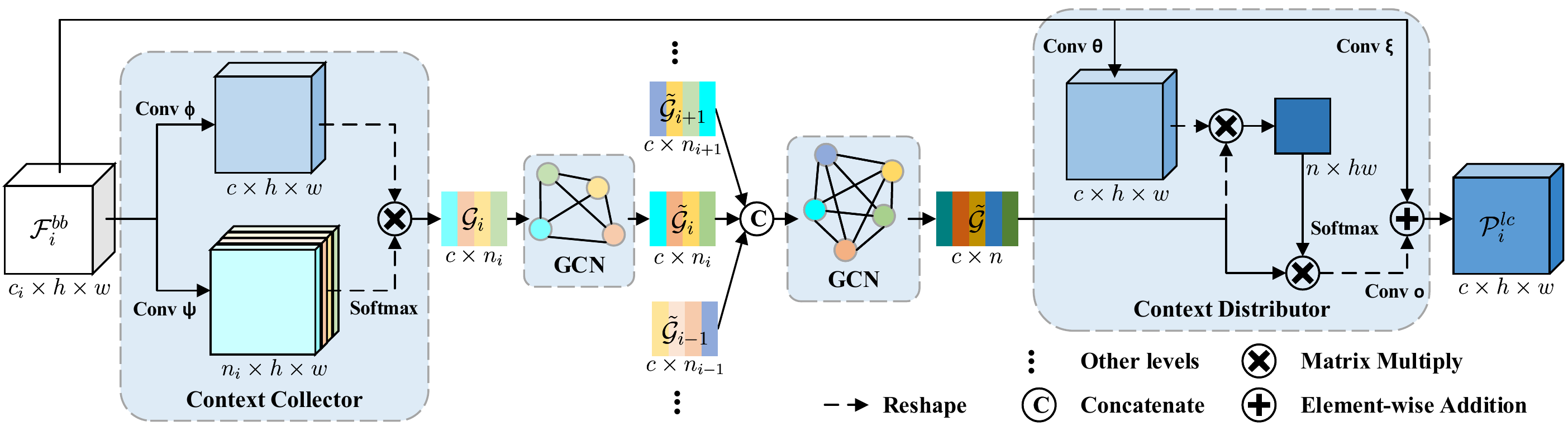}
	\caption{\small{\textbf{Detailed illustration of the Multi-level Global Context (MGC) module.} MGC collects, reasons, and distributes multi-level global context features through three children modules, the Context Collector, the GCNs, and the Context Distributor, respectively.}}
	\label{fig: illustration of MGC}
	\vspace{-10pt}
\end{figure*}

\subsection{Multi-level Global Context}
\label{subsection: multi-level global context}
In feature extraction, pyramidal features suffer from information loss due to channel reduction. ParseNet \cite{liu2015parsenet} distributes the global context feature to all locations for information supplement. However, a single context feature collected by global average pooling ignores different needs across scales and locations. Inspired by this, we propose the Multi-level Global Context (MGC) module to extract more discriminative features by aggregating multi-level global context features and mitigate semantic information loss.

MGC consists of three steps to adaptively aggregate global context features from the feature hierarchy, as shown in Figure \ref{fig: illustration of MGC}. First, the Context Collector collects global context features from all feature levels through attention pooling. Second, the Graph Convolutional Networks (GCNs) are employed to reason contextual relations. Third, the Context Distributor distributes the context features to each level. Each step will be discussed in detail as follows.

\textbf{Context Collector.} 
Given the i-th feature $\mathcal{F}^{bb}_{i} \in \mathbb{R}^{c_{i} \times h \times w}$ from the backbone, where $c_{i}$ denotes channel dimension and $h, w$ are spatial size, Context Collector generates a new feature $ \mathcal{G}_{i} \in \mathbb{R}^{c \times n_i} $ comprising $n_i$ different context features of dimension $c$. Inspired by AN \cite{wang2020attentive}, we assume that the features of different stages differ in semantic richness and intensity, and are composed of diverse semantic entities. Here the feature $\mathcal{F}^{bb}_{i}$ is supposed to consist of $n_i$ semantic entities, and the feature points are gathered into different context features according to their cosine similarity to the semantic entities.

In Context Collector, one $1 \times 1$ convolutional layer $\psi$ with $n_i$ filters is adopted as semantic entities, and another $1 \times 1$ convolutional layer $\phi$ is used to embed the input feature. In particular, we formulate this procedure as Eqn. \ref{eq: context collector}.
\begin{equation}
	\label{eq: context collector}
	\mathcal{G}_i = W_{\phi} \mathcal{F}^{bb}_{i} \mathrm{Softmax} (\sqrt{c_{i}} \cdot W_{\psi} \mathrm{Norm}(\mathcal{F}^{bb}_{i}) ) ^T,
\end{equation}
\begin{equation}
	\label{eq: orthogonal regularization}
	L_o = \lambda_o \|W_{\psi}W_{\psi}^T - I\|^2_F,
\end{equation}
where $ W_{\psi} \in \mathbb{R}^{n_i \times c_{i}} $, $ W_{\phi} \in \mathbb{R}^{c \times c_{i}} $, and $\mathrm{Norm}$ represents $L_2$ normalization. To normalize the semantic entities for computing cosine similarity and learn more diverse patterns, orthogonal regularization is applied to the weights $W_{\psi}$ as shown in Eqn. \ref{eq: orthogonal regularization}. Different from the scaled dot-product attention in \cite{vaswani2017attention} which divides each dot product by $\sqrt{c_{i}}$, we multiply the cosine similarity by $\sqrt{c_{i}}$ to maintain the variance at different values of $c_{i}$. Then we apply a softmax function spatially to generate the attention masks, and collect context features using attention pooling accordingly.

We call our particular attention mechanism ``Scaled Cosine-similarity Attention'', which computes the compatibility function based on scaled cosine similarity. It focuses on semantic content instead of intensity and can avoid strongly activated keys surpassing other keys. For brevity, we reformulate the compatibility function out of Eqn. \ref{eq: context collector}:
\begin{equation}
	\label{eq: scaled cosine-similarity attention}
	\boldsymbol{CF} (W^{T}_{\psi}, \mathcal{F}^{bb}_{i}, c_i) = \mathrm{Softmax}(\sqrt{c_{i}} \cdot W_{\psi} \mathrm{Norm}(\mathcal{F}^{bb}_{i}) ) ^T.
\end{equation}

\textbf{GCN.} 
Once we obtain context features $ \mathcal{G}_i \in \mathbb{R}^{c \times n_i} $, we utilize GCNs to model and reason the contextual relations between them first in single level. The formula of the GCN with a residual connection \cite{he2016deep} is interpreted as Eqn. \ref{eq: gcn single level}, where $W^{gcn}_{1} \in \mathbb{R}^{\frac{c}{4} \times c}, W^{gcn}_{2} \in \mathbb{R}^{\frac{c}{4} \times c}, W^{gcn}_{3} \in \mathbb{R}^{c \times c}$ are weight matrices of 1D convolutional layers. The convolutional weight of the GCN is $ W^{gcn}_{3} $ and the adjacent matrix is dynamically generated through self-attention described in Eqn. \ref{eq: scaled cosine-similarity attention}. After contextual relations are reasoned in each level independently, all context features are concatenated and reasoned together in multi-level by the GCN as Eqn. \ref{eq: gcn multi-level}. In this work, we collect $\{ n_2, n_3, n_4, n_5 \}$ context features from the feature maps $\{\mathcal{F}^{bb}_{2}, \mathcal{F}^{bb}_{3}, \mathcal{F}^{bb}_{4}, \mathcal{F}^{bb}_{5}\}$, respectively.
\begin{equation}
	\label{eq: gcn single level}
	\tilde{\mathcal{G}}_i = W^{gcn}_{3} \mathcal{G}_i \boldsymbol{CF} (W^{gcn}_{1} \mathcal{G}_i, W^{gcn}_{2} \mathcal{G}_i, \frac{c}{4}) + \mathcal{G}_i,
\end{equation}
\begin{equation}
	\label{eq: gcn multi-level}
	\tilde{\mathcal{G}} = \boldsymbol{GCN} ([\cdots, \tilde{\mathcal{G}}_{i-1}, \tilde{\mathcal{G}}_{i}, \tilde{\mathcal{G}}_{i+1}, \cdots]).
\end{equation}

\textbf{Context Distributor.} 
Context Distributor distributes the multi-level global context features $\tilde{\mathcal{G}} \in \mathbb{R}^{c \times n}$ to each level using the proposed scaled cosine-similarity attention. As shown in Eqn. \ref{eq: context distributor}, $W_{o} \in \mathbb{R}^{c \times c}$, $W_{\theta} \in \mathbb{R}^{c \times c_i}$ and $W_{\xi} \in \mathbb{R}^{c \times c_i}$ are weights of $1 \times 1$ convolutional layers. The feature maps $\mathcal{F}^{bb}_{i}$ are queries and the global context features $\tilde{\mathcal{G}}$ are shared as keys and values. 
\begin{equation}
	\label{eq: context distributor}
	\mathcal{P}^{lc}_{i} = W_{o} \tilde{\mathcal{G}} \boldsymbol{CF} (W_{\theta}\mathcal{F}^{bb}_{i}, \tilde{\mathcal{G}}, c) + W_{\xi}\mathcal{F}^{bb}_{i}.
\end{equation}

\begin{figure*}[htbp]
	\centering
	\includegraphics[width=1.0\linewidth]{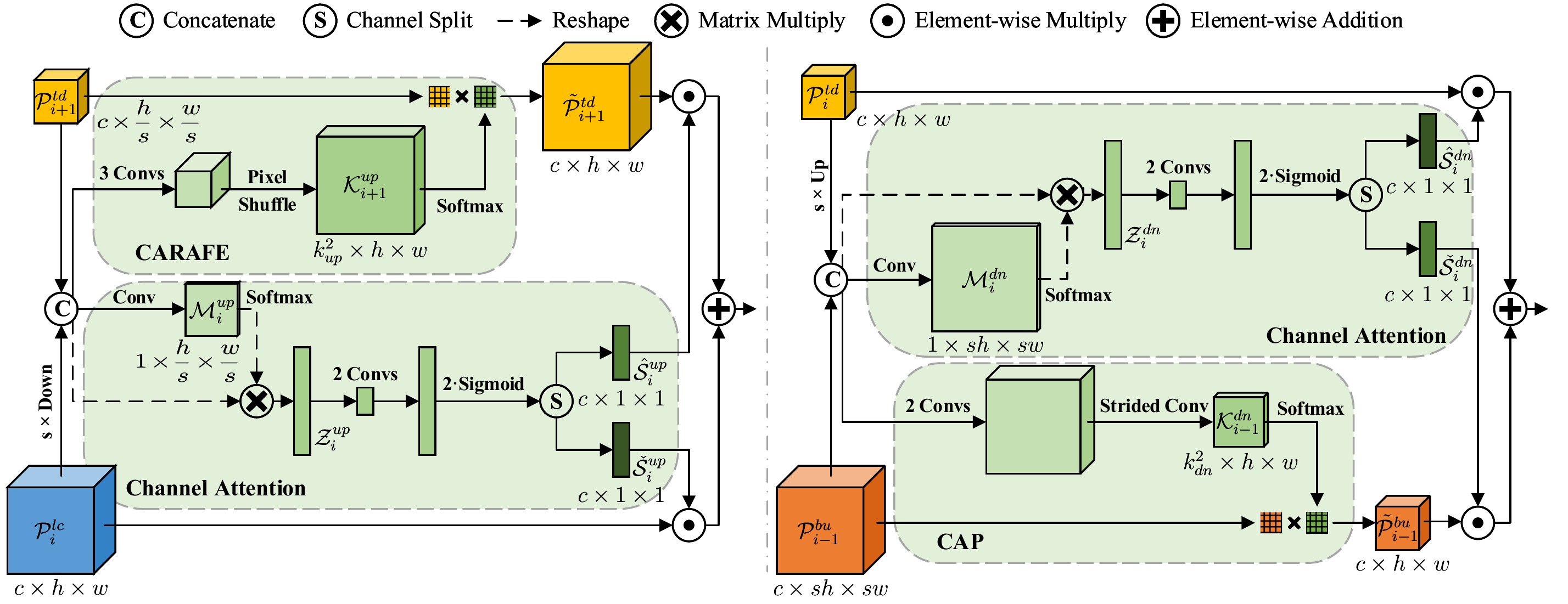}
	\caption{\small{\textbf{The details of the Global Attention CARAFE (GACARAFE) module (left) and the Global Attention Content-Aware Pooling (GACAP) module (right).} Noting that the fused features $\mathcal{P}^{td}_{i}$ and $\mathcal{P}^{bu}_{i}$ after element-wise addition are omitted for simplification.}}
	\label{fig: illustration of GACARAFE and GACAP}
	\vspace{-10pt}
\end{figure*}

\subsection{Global Attention CARAFE}
\label{subsection: global attention carafe}
In the top-down pathway of feature fusion, interpolation depends only on the relative positions and direct element-wise addition ignores the semantic gap between adjacent features. Recently, CARAFE \cite{wang2019carafe} is proposed to upsample features through content-aware reassembly. However, CARAFE only leverages the information of low-resolution feature and fails to capture complementary semantics from the high-resolution level. As shown in Figure \ref{fig: illustration of GACARAFE and GACAP} (left), we propose the Global Attention CARAFE (GACARAFE) module for feature fusion, where the complementary information is aggregated to guide content-aware upsampling and channel attention \cite{hu2018squeeze} is employed to enhance the semantic consistency before merging adjacent features.

The feature $\mathcal{P}^{lc}_{i} \in \mathbb{R}^{c \times h \times w}$ is extracted through the MGC module and the feature $\mathcal{P}^{td}_{i+1} \in \mathbb{R}^{c \times \frac{h}{s} \times \frac{w}{s}}$ is the upper level in the top-down pathway, where $s=2$ is the scale factor. We downsample the feature $ \mathcal{P}^{lc}_{i} $ using max-pooling and concatenate $\mathcal{P}^{td}_{i+1}$ and $\tilde{\mathcal{P}}^{lc}_{i}$ together. As shown in Figure \ref{fig: illustration of GACARAFE and GACAP} (left), the feature $ \mathcal{P}^{td}_{i+1} $ is upsampled as $ \tilde{\mathcal{P}}^{td}_{i+1} $ through two steps. GACARAFE predicts a location-wise kernel $\mathcal{K}^{up}_{i+1}(x, y)$ based on the $k^{up}_{en} \times k^{up}_{en}$ neighbor of feature points $ \mathcal{P}^{td}_{i+1}(\frac{x}{s}, \frac{y}{s})$ and $ \tilde{\mathcal{P}}^{lc}_{i}(\frac{x}{s}, \frac{y}{s})$, as shown in Eqn. \ref{eq: gacarafe kernel}. And then it reassembles the $k_{up} \times k_{up}$ neighbor of $ \mathcal{P}^{td}_{i+1}(\frac{x}{s}, \frac{y}{s})$ with the kernel  $\mathcal{K}^{up}_{i+1}(x, y) $, as presented in Eqn. \ref{eq: gacarafe reassembly}.
\begin{equation}
	\label{eq: gacarafe kernel}
	\mathcal{K}^{up}_{i+1}(x, y) = \boldsymbol{K} (\mathcal{N}([\mathcal{P}^{td}_{i+1}, \tilde{\mathcal{P}}^{lc}_{i}](\frac{x}{s}, \frac{y}{s}), k^{up}_{en})),
\end{equation}
\begin{equation}
	\label{eq: gacarafe reassembly}
	\tilde{\mathcal{P}}^{td}_{i+1}(x, y) = \boldsymbol{R} (\mathcal{N}(\mathcal{P}^{td}_{i+1}(\frac{x}{s}, \frac{y}{s}), k_{up}), \mathcal{K}^{up}_{i+1}(x, y)),
\end{equation}
where $ \boldsymbol{K} $ and $ \boldsymbol{R} $ are the kernel prediction module and content-aware reassembly module altered from CARAFE \cite{wang2019carafe}, respectively. The kernel prediction module contains three convolutions: the first one of $1 \times 1$ kernel for reducing channel dimension from $2c$ to $c^{up}_{m}$, the second one of $3 \times 3$ kernel with ReLU activation for blending features, and the third one of $k^{up}_{en} \times k^{up}_{en}$ kernel for predicting upsampling kernels. The kernels $\mathcal{K}^{up}_{i+1}$ are deformed using pixel shuffle \cite{shi2016real} and normalized with a softmax function.

Besides, we apply channel attention to learn two vectors $\hat{\mathcal{S}}^{up}_{i}$ and $\check{\mathcal{S}}^{up}_{i}$ to recalibrate the adjacent features. As shown in Eqn. \ref{eq: gacarafe channel attention}, we squeeze the global spatial information into a channel descriptor using attention pooling. Specifically, a $1 \times 1$ convolutional layer is employed to predict the attention mask $\mathcal{M}^{up}_{i}$ and channel-wise statistics $\mathcal{Z}^{up}_{i}$ is generated by shrinking the concatenated feature accordingly. And then a simple gating mechanism with $2 \cdot \mathrm{sigmoid}$ activation is adopted to model channel-wise interdependencies.
\begin{equation}
	\label{eq: gacarafe channel attention}
	\begin{split}
		\mathcal{M}^{up}_{i} = \mathrm{Softmax} & (W^{up}_{1} [\mathcal{P}^{td}_{i+1}, \tilde{\mathcal{P}}^{lc}_{i}])^T, \\
		\mathcal{Z}^{up}_{i} = [\mathcal{P}^{td}_{i+1}, & \tilde{\mathcal{P}}^{lc}_{i}] \mathcal{M}^{up}_{i}, \\
		[\hat{\mathcal{S}}^{up}_{i}, \check{\mathcal{S}}^{up}_{i}] = 2 \sigma (W^{up}_{3} & \mathrm{ReLU}(\mathrm{LN}(W^{up}_{2}\mathcal{Z}^{up}_{i}))).
	\end{split}
\end{equation}
Here $\mathrm{LN}$ stands for LayerNorm \cite{ba2016layer}, $W^{up}_{1} \in \mathbb{R}^{1 \times 2c}$, $W^{up}_{2} \in \mathbb{R}^{\frac{c}{2} \times 2c} $ and $W^{up}_{3} \in \mathbb{R}^{2c \times \frac{c}{2}} $. $2\sigma$ indicates the activation function $2 \cdot \mathrm{sigmoid}$, which can keep the mean of channel-wise weights being 1 after successive multiplications and excite or restrain features selectively. Finally, the adjacent features $ \tilde{\mathcal{P}}^{td}_{i+1} $ and $ \mathcal{P}^{lc}_{i} $ are reweighted and merged by element-wise addition as shown in Eqn. \ref{eq: gacarafe addition}.
\begin{equation}
	\label{eq: gacarafe addition}
	\mathcal{P}^{td}_{i} = \hat{\mathcal{S}}^{up}_{i} \odot \tilde{\mathcal{P}}^{td}_{i+1} + \check{\mathcal{S}}^{up}_{i} \odot \mathcal{P}^{lc}_{i}.
\end{equation}

\subsection{Global Attention Content-Aware Pooling}
\label{subsection: global attention content-aware pooling}
In the bottom-up pathway, strided convolution applies the same kernel across the entire image and ignores the underlying content. To aggregate more discriminative features, we propose the Global Attention Content-Aware Pooling (GACAP) module to execute the bottom-up fusion. As shown in Figure \ref{fig: illustration of GACARAFE and GACAP} (right), the Content-Aware Pooling (CAP) operator extends the idea of CARAFE \cite{wang2019carafe} to feature pooling and the GACAP module differs from the GACARAFE module in several aspects.

First, GACAP unsamples the feature $\mathcal{P}^{td}_{i} \in \mathbb{R}^{c \times h \times w}$ to get $ \tilde{\mathcal{P}}^{td}_{i} $ through bilinear interpolation and concatenates $\mathcal{P}^{bu}_{i-1} \in \mathbb{R}^{c \times sh \times sw}$ with it. Second, GACAP downsamples the feature $\mathcal{P}^{bu}_{i-1}$ as $\tilde{\mathcal{P}}^{bu}_{i-1}$, as shown in Eqn. \ref{eq: gacap kernel} and \ref{eq: gacap reassembly}.
\begin{equation}
	\label{eq: gacap kernel}
	\mathcal{K}^{dn}_{i-1}(x, y) = \boldsymbol{K} (\mathcal{N}([\mathcal{P}^{bu}_{i-1}, \tilde{\mathcal{P}}^{td}_{i}](sx, sy), k^{dn}_{en})),
\end{equation}
\begin{equation}
	\label{eq: gacap reassembly}
	\tilde{\mathcal{P}}^{bu}_{i-1}(x, y) = \boldsymbol{R} (\mathcal{N}(\mathcal{P}^{bu}_{i-1}(sx, sy), k_{dn}), \mathcal{K}^{dn}_{i-1}(x, y)),
\end{equation}
where the kernel prediction module $ \boldsymbol{K} $ applies a $k^{dn}_{en} \times k^{dn}_{en}$ convolutional layer with stride $s$ to generate downsampling kernels $\mathcal{K}^{dn}_{i-1}$ directly. The other parts of GACAP is exactly the same with GACARAFE described in Section \ref{subsection: global attention carafe}. After the channel attention vectors $\hat{\mathcal{S}}^{dn}_{i}$ and $\check{\mathcal{S}}^{dn}_{i}$ are obtained, the adjacent features are fused to generate feature $\mathcal{P}^{bu}_{i}$ as Eqn. \ref{eq: gacap addition}. Notably, we finally append a $3 \times 3$ convolution on each merged feature map to reduce the aliasing effect, whether in the top-down pathway or the bottom-up pathway.
\begin{equation}
	\label{eq: gacap addition}
	\mathcal{P}^{bu}_{i} = \hat{\mathcal{S}}^{dn}_{i} \odot \mathcal{P}^{td}_{i} + \check{\mathcal{S}}^{dn}_{i} \odot \tilde{\mathcal{P}}^{bu}_{i-1}.
\end{equation}

\section{Experiments}
\label{section: experiments}
\subsection{Datasets and Evaluation Metrics}
\label{subsection: datasets and evaluation metrics}
\textbf{Datasets.} We conduct experiments on the challenging MS COCO dataset \cite{lin2014microsoft}. It contains 115k images for training (\emph{train2017}), 5k images for validation (\emph{val2017}), and 20k images for testing (\emph{test-dev}). The labels of \emph{test-dev} are not publicly available. We train our models on \emph{train2017} subset and report results on \emph{val2017} subset for ablation study. we also report results on \emph{test-dev} for comparison.

\textbf{Evaluation Metrics.} We report the standard COCO-style Average Precision (AP) metric including AP (averaged over IoU thresholds), $\mathrm{AP}_{50}$, $\mathrm{AP}_{75}$ (AP at different IoU thresholds), and $\mathrm{AP}_{S}$, $\mathrm{AP}_{M}$, $\mathrm{AP}_{L}$ (AP at different scales). Since our framework is general to both instance segmentation and object detection, both mask AP and box AP (superscripted as ``$bb$'') are evaluated.

\subsection{Implementation Details}
\label{subsection: implementation details}
All experiments are implemented based on MMDetection \cite{mmdetection}. We train detectors with a batchsize of 8 over 4 GPUs (2 images per GPU) for 12 epochs with an initial learning rate of 0.01, and decrease it by 0.1 after 8 and 11 epochs, respectively. Images are resized to a maximum scale of $1333 \times 800$ pixels without changing the aspect ratio. If not otherwise specified, $A^2$-FPN adopts a fixed set of hyper-parameters in experiments, where $c=256$, $n_i = 64(6 - i)$, and $\lambda_o = 0.0001$ in Eqn. \ref{eq: orthogonal regularization} in the MGC module, $C^{up}_{m}=64, k^{up}_{en}=3$ and $ k_{up}=5$ in the GACARAFE module, $C^{dn}_{m}=64, k^{dn}_{en}=3$ and $k_{dn}=5$ in the GACAP module. All other hyper-parameters in this work follow the default setting of MMDetection \cite{mmdetection}.

\def\compimgheight{2.3cm}

\begin{figure*}[htbp]
	\centering
	\includegraphics[height=\compimgheight]{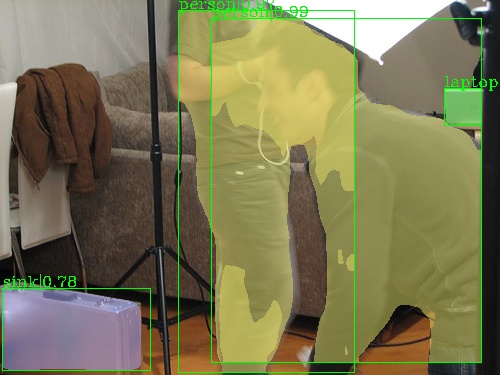}
	\includegraphics[height=\compimgheight]{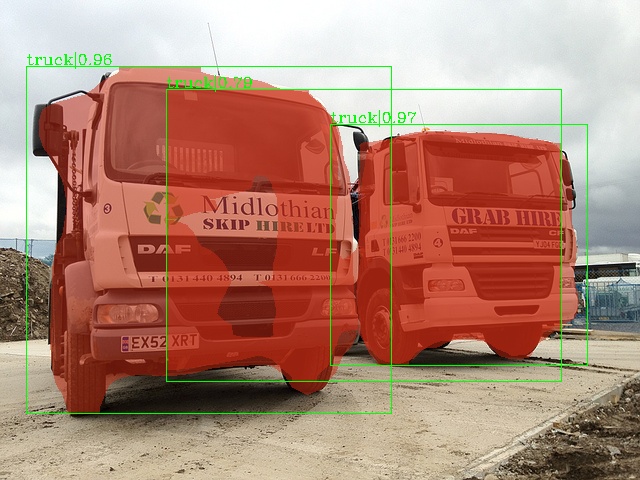}
	\includegraphics[height=\compimgheight]{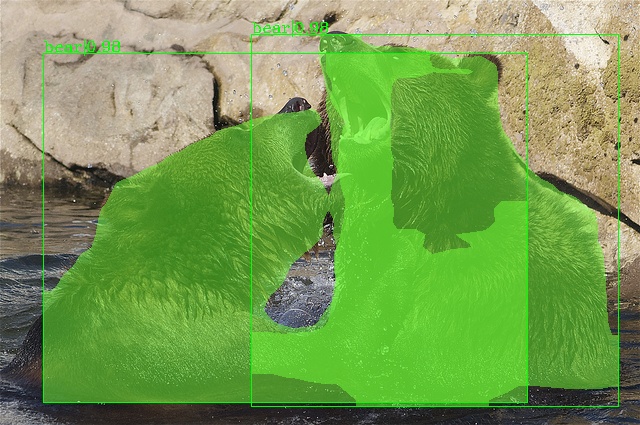}
	\includegraphics[height=\compimgheight]{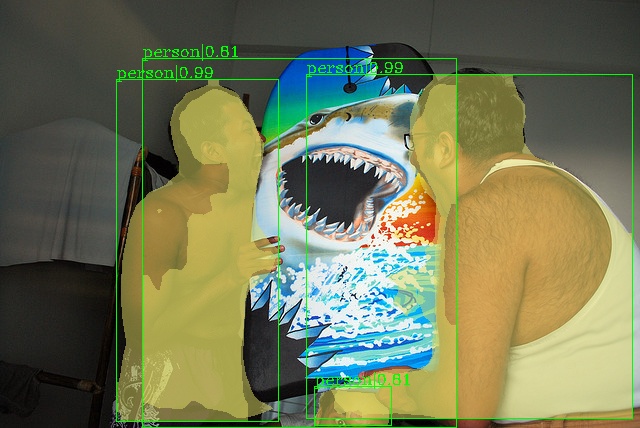}
	\includegraphics[height=\compimgheight]{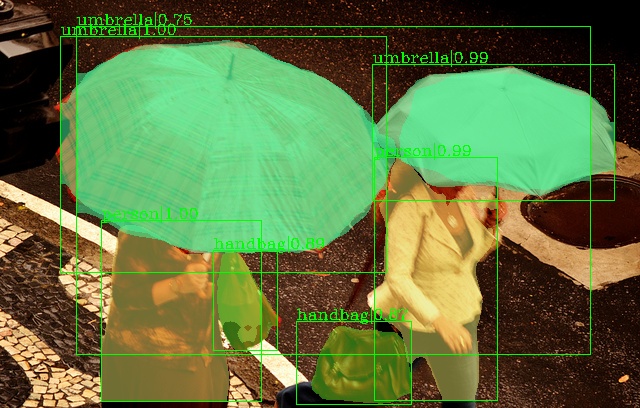}
	
	\includegraphics[height=\compimgheight]{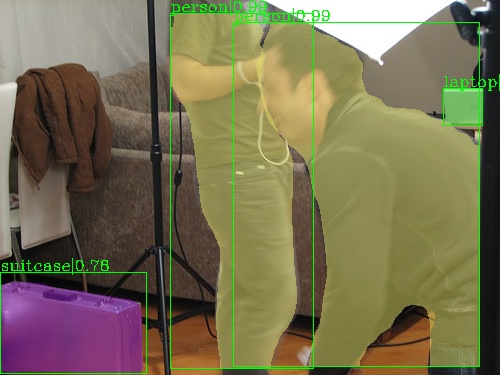}
	\includegraphics[height=\compimgheight]{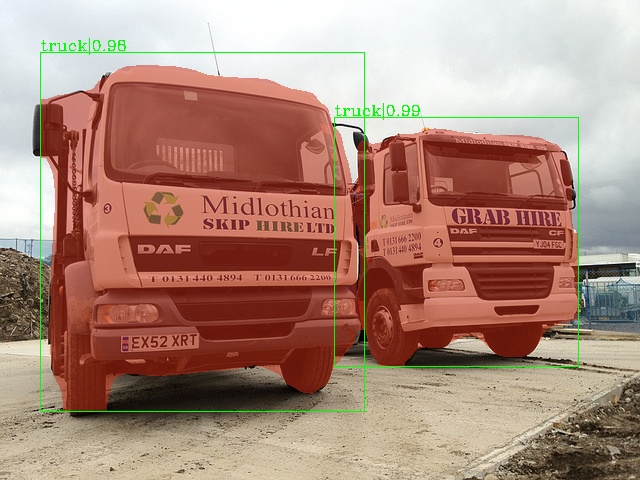}
	\includegraphics[height=\compimgheight]{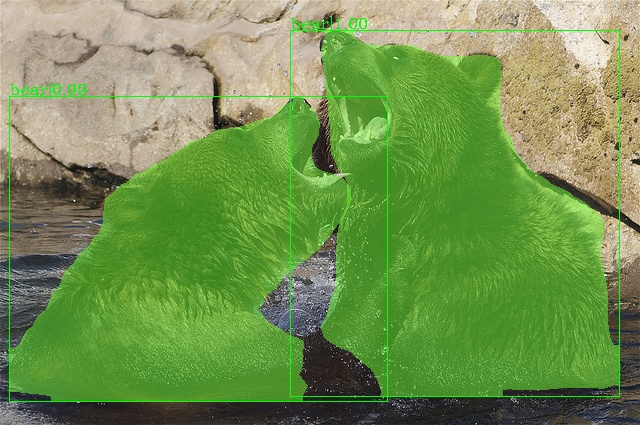}
	\includegraphics[height=\compimgheight]{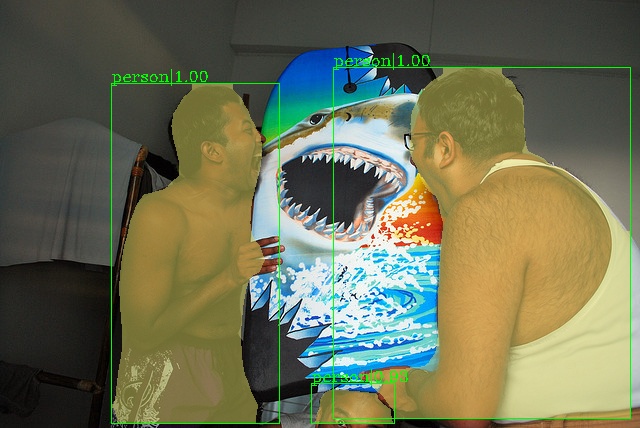}
	\includegraphics[height=\compimgheight]{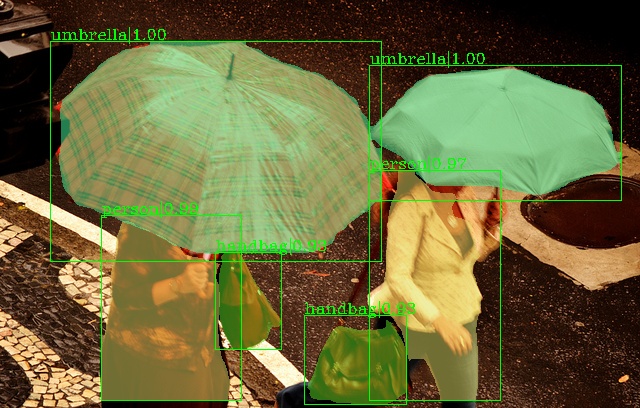}
	\caption{\small{Comparison of instance segmentation results between \textbf{FPN} (top row) and $\boldsymbol{A^2}$\textbf{-FPN} (bottom row) on COCO \emph{val2017}.}}
	\label{fig: comparison}
\end{figure*}

\begin{table*}[htbp]
	\centering
	\caption{\small{\textbf{Instance segmentation mask AP and object detection box AP}, vs. the state-of-the-art on COCO \textit{test-dev}. ``\textit{ms}'' in [] indicates multi-scale training and the symbol `*' denotes our re-implementation results. The letters 'R' and 'X' stand for the backbone networks ResNet \cite{he2016deep} and ResNeXt \cite{xie2017aggregated}, respectively. `Sch.' is short for the training schedule, which follows the setting of MMdetection \cite{mmdetection}.}}
	\resizebox{1.0\textwidth}{!}{
		\begin{tabular}{ccc|cccccc|cccccc}
			\toprule
			Method & Backbone & Sch. & AP & $\mathrm{AP}_{50}$ & $\mathrm{AP}_{75}$ & $\mathrm{AP}_{S}$ & $\mathrm{AP}_{M}$ & $\mathrm{AP}_{L}$ & $\mathrm{AP}^{bb}$ & $\mathrm{AP}^{bb}_{50}$ & $\mathrm{AP}^{bb}_{75}$ & $\mathrm{AP}^{bb}_{S}$ & $\mathrm{AP}^{bb}_{M}$ & $\mathrm{AP}^{bb}_{L}$ \\
			\midrule
			\midrule
			Faster R-CNN \cite{lin2017feature} & R-101-FPN & - & - & - & - & - & - & - & 36.2 & 59.1 & - & 18.2 & 39.0 & 48.2 \\
			Mask R-CNN \cite{he2017mask} & R-101-FPN & - & 35.7 & 58.0 & 37.8 & 15.5 & 38.1 & 52.4 & 38.2 & 60.3 & 41.7 & 20.1 & 41.1 & 50.2 \\
			Mask R-CNN \cite{he2017mask} & X-101-FPN & - & 37.1 & 60.0 & 39.4 & 16.9 & 39.9 & 53.5 & 39.8 & 62.3 & 43.4 & 22.1 & 43.2 & 51.2 \\
			Mask R-CNN \cite{guo2020augfpn} & R-101-AugFPN & 1x & 37.8 & 60.4 & 40.4 & 20.4 & 41.0 & 49.8 & 41.3 & 63.5 & 44.9 & 24.2 & 44.8 & 52.0 \\
			PANet \cite{liu2018path} & R-50-PAFPN & - & 36.6 & 58.0 & 39.3 & 16.3 & 38.1 & 53.1 & 41.2 & 60.4 & 44.4 & 22.7 & 44.0 & 54.6 \\
			PANet[\textit{ms}] \cite{liu2018path} & R-50-PAFPN & - & 38.2 & 60.2 & 41.4 & 19.1 & 41.1 & 52.6 & 42.5 & 62.3 & 46.4 & 26.3 & 47.0 & 52.3 \\
			PANet \cite{liu2018path} & X-101-PAFPN & - & 40.0 & 62.8 & 43.1 & 18.8 & 42.3 & 57.2 & 45.0 & 65.0 & 48.6 & 25.4 & 48.6 & 59.1 \\
			PANet[\textit{ms}] \cite{liu2018path} & X-101-PAFPN & - & 42.0 & 65.1 & 45.7 & 22.4 & 44.7 & 58.1 & 47.4 & 67.2 & 51.8 & 30.1 & 51.7 & 60.0 \\
			Cascade R-CNN \cite{cai2018cascade} & R-50-FPN & - & - & - & - & - & - & - & 40.6 & 59.9 & 44.0 & 22.6 & 42.7 & 52.1 \\
			Cascade R-CNN \cite{cai2018cascade} & R-101-FPN & - & - & - & - & - & - & - & 42.8 & 62.1 & 46.3 & 23.7 & 45.5 & 55.2 \\
			HTC \cite{chen2019hybrid} & R-50-FPN & 20e & 38.4 & 60.0 & 41.5 & 20.4 & 40.7 & 51.2 & 43.6 & - & - & - & - & - \\
			HTC \cite{chen2019hybrid} & R-101-FPN & 20e & 39.7 & 61.8 & 43.1 & 21.0 & 42.2 & 53.5 & 45.3 & - & - & - & - & - \\
			HTC \cite{chen2019hybrid} & X-101-FPN & 20e & 41.2 & 63.9 & 44.7 & 22.8 & 43.9 & 54.6 & 47.1 & - & - & - & - & - \\
			\midrule
			\midrule
			Mask R-CNN* & R-50-FPN & 1x & 34.5 & 56.3 & 36.7 & 18.6 & 37.3 & 44.7 & 37.6 & 59.5 & 40.6 & 21.8 & 40.8 & 46.4 \\
			Mask R-CNN(ours) & R-50-$A^2$-FPN-Lite & 1x & \textbf{36.4} & \textbf{58.9} & \textbf{38.7} & \textbf{19.6} & \textbf{39.1} & \textbf{47.8} & \textbf{39.8} & \textbf{62.3} & \textbf{43.4} & \textbf{23.6} & \textbf{42.8} & \textbf{49.9} \\
			Mask R-CNN(ours) & R-50-$A^2$-FPN & 1x & \textbf{36.6} & \textbf{59.3} & \textbf{39.1} & \textbf{19.8} & \textbf{39.3} & \textbf{48.0} & \textbf{40.2} & \textbf{62.7} & \textbf{43.7} & \textbf{23.7} & \textbf{43.2} & \textbf{50.2} \\
			Mask R-CNN[\textit{ms}](ours) & R-50-$A^2$-FPN & 2x & \textbf{38.8} & \textbf{62.0} & \textbf{41.6} & \textbf{22.1} & \textbf{41.7} & \textbf{50.4} & \textbf{42.8} & \textbf{65.2} & \textbf{47.0} & \textbf{26.5} & \textbf{45.9} & \textbf{53.3} \\
			\midrule
			Mask R-CNN* & R-101-FPN & 1x & 36.3 & 58.5 & 38.9 & 19.4 & 39.3 & 47.8 & 39.9 & 61.6 & 43.6 & 23.1 & 43.2 & 50.0 \\
			Mask R-CNN(ours) & R-101-$A^2$-FPN & 1x & \textbf{37.9} & \textbf{60.8} & \textbf{40.5}  & \textbf{20.6} & \textbf{41.8} & \textbf{50.1} & \textbf{41.7} & \textbf{64.1} & \textbf{45.5} & \textbf{24.6} & \textbf{45.0} & \textbf{52.5} \\
			\midrule
			Cascade Mask R-CNN* & R-50-FPN & 1x & 36.1 & 57.1 & 38.9 & 19.1 & 38.6 & 47.4 & 41.5 & 59.8 & 45.1 & 23.4 & 44.3 & 52.7 \\
			Cascade Mask R-CNN(ours) & R-50-$A^2$-FPN & 1x & \textbf{38.1} & \textbf{60.1} & \textbf{41.0} & \textbf{20.6} & \textbf{40.5} & \textbf{50.4} & \textbf{43.9} & \textbf{62.8} & \textbf{47.7} & \textbf{25.4} & \textbf{46.8} & \textbf{56.0} \\
			\midrule
			Cascade Mask R-CNN* & R-101-FPN & 1x & 37.4 & 58.9 & 40.5 & 19.5 & 40.1 & 49.6 & 43.3 & 61.6 & 47.2 & 24.1 & 46.2 & 55.3 \\
			Cascade Mask R-CNN(ours) & R-101-$A^2$-FPN & 1x & \textbf{39.1} & \textbf{61.3} & \textbf{42.2}  & \textbf{21.0} & \textbf{41.9} & \textbf{51.8} & \textbf{45.1} & \textbf{64.1} & \textbf{48.9} & \textbf{25.7} & \textbf{48.2} & \textbf{57.7} \\
			\midrule
			HTC* & R-50-FPN & 20e & 38.4 & 60.0 & 41.4 & 20.3 & 40.6 & 51.2 & 43.5 & 62.6 & 47.3 & 24.5 & 45.9 & 55.9 \\
			HTC(ours) & R-50-$A^2$-FPN & 20e & \textbf{39.8} & \textbf{62.3} & \textbf{43.0} & \textbf{21.6} & \textbf{42.4} & \textbf{52.8} & \textbf{45.4} & \textbf{64.9} & \textbf{49.1} & \textbf{26.3} & \textbf{48.2} & \textbf{57.7} \\
			\midrule
			HTC* & R-101-FPN & 20e & 39.6 & 61.6 & 42.9 & 21.1 & 42.2 & 53.1 & 45.1 & 64.3 & 49.0 & 25.7 & 47.9 & 58.2 \\
			HTC(ours) & R-101-$A^2$-FPN & 20e & \textbf{40.8} & \textbf{63.6} & \textbf{44.1} & \textbf{22.3} & \textbf{43.5} & \textbf{54.4} & \textbf{46.6} & \textbf{66.2} & \textbf{50.4} & \textbf{27.1} & \textbf{49.6} & \textbf{59.9} \\
			\midrule
			HTC* & X-101-FPN & 20e & 41.3 & 64.0 & 44.8 & 22.7 & 43.9 & 54.8 & 47.2 & 66.6 & 51.4 & 27.7 & 50.0 & 60.2 \\
			HTC(ours) & X-101-$A^2$-FPN & 20e & \textbf{42.1} & \textbf{65.3} & \textbf{45.7} & \textbf{23.6} & \textbf{44.8} & \textbf{56.0} & \textbf{48.3} & \textbf{68.0} & \textbf{52.4} & \textbf{28.9} & \textbf{51.3} & \textbf{61.7} \\
			HTC[\textit{ms}](ours) & X-101-$A^2$-FPN & 2x & \textbf{44.0} & \textbf{67.5} & \textbf{47.9} & \textbf{25.7} & \textbf{47.1} & \textbf{57.6} & \textbf{50.4} & \textbf{70.1} & \textbf{54.9} & \textbf{31.6} & \textbf{53.7} & \textbf{63.7} \\
			\bottomrule
		\end{tabular}
	}
	\label{tab: instance segmentation and object detection results}
	\vspace{-10pt}
\end{table*}

\subsection{Benchmarking Results}
\label{subsection: benchmarking results}
We integrate $A^2$-FPN into the state-of-the-art detectors and evaluate the performance on the COCO \emph{test-dev}. For a fair comparison, we re-implement the corresponding FPN based methods as baselines.

\textbf{Instance Segmentation Results.} $A^2$-FPN, combined with different instance segmentation frameworks and backbone networks, achieves better performance compared with baseline models. As shown in Table \ref{tab: instance segmentation and object detection results}, by replacing FPN with $A^2$-FPN, Mask R-CNN using ResNet-50 (denoted as R-50-$A^2$-FPN) achieves 36.6\% mask AP, which is 2.1\% higher than the counterpart. Moreover, when using ResNet-101 as the feature extractor, $A^2$-FPN still brings a gain of 1.6\%, proving that $A^2$-FPN can consistently improve the performance even with a more powerful backbone. 

$A^2$-FPN-Lite is the lightweight setting of $A^2$-FPN including three differences, halving the channel ($c = 128$) and number ($n_i = 32(6 - i)$) of context features, removing $\mathcal{F}^{bb}_{6}$ and generating $\mathcal{F}^{bu}_{6}$ from $\mathcal{F}^{bu}_{5}$ using max-pooling like FPN \cite{lin2017feature}, and removing the $3 \times 3$ convolution after $\mathcal{F}^{bu}_{2}$ like PAFPN \cite{liu2018path}. As shown in Table \ref{tab: instance segmentation and object detection results}, $A^2$-FPN-Lite improves the performance of Mask R-CNN by 1.9\% mask AP. 

Besides, our proposed method works well with different frameworks. $A^2$-FPN combined with Cascade Mask R-CNN leads to a gain of 2.0\% and 1.7\% for ResNet-50 and ResNet-101, respectively. It is noteworthy that $A^2$-FPN further boosts the performance of HTC by 1.4\% when using ResNet-50 as backbone, while PAFPN only shows a limited 0.1\% gain in the extensive study in HTC \cite{chen2019hybrid}. Meanwhile, HTC achieves 40.8\% and 42.1\% mask AP when using R-101-$A^2$-FPN and X-101-$A^2$-FPN as backbone, pushing the powerful HTC by 1.2\% and 0.8\%, respectively. Particularly, $A^2$-FPN equipped with HTC achieves 44.0\% mask AP under the setting of a `2x' schedule and multi-scale training. The improvements demonstrate that $A^2$-FPN further improves the multi-scale feature learning of feature pyramid through attention-guided aggregation. In Figure \ref{fig: comparison}, we show some examples of instance segmentation results, where $A^2$-FPN generates more accurate instance masks compared to the FPN based baseline.

\textbf{Object Detection Results.} We also evaluate $A^2$-FPN on object detection task as shown in Table \ref{tab: instance segmentation and object detection results}. By replacing FPN with $A^2$-FPN in Mask R-CNN, the performance is boosted by 2.6\% and 1.8\% for ResNet-50 and ResNet-101. $A^2$-FPN-Lite also boosts the performance of Mask R-CNN by 2.2\%. Moreover, Cascade Mask R-CNN can be improved by 2.4\% and 1.8\% correspondingly when using ResNet-50 and ResNet-101 as backbone. Meanwhile, $A^2$-FPN equipped with HTC contributes a gain of 1.9\%, 1.5\%, and 1.1\% for ResNet-50, ResNet-101, and ResNeXt-101, respectively. When adopting the setting of a `2x' schedule and multi-scale training, $A^2$-FPN achieves 50.4\% box AP. In brief, $A^2$-FPN effectively improves the performance of state-of-the-art detectors not only on instance segmentation but also on object detection. As shown in Table \ref{tab: instance segmentation and object detection results}, $A^2$-FPN brings consistent gains on various backbone networks, frameworks, and even different tasks, which verifies the effectiveness and generalization ability of $A^2$-FPN.

\begin{table}[tp]
	\centering
	\caption{\small{\textbf{Effect of each module in our design.} MC: the MGC module, GE: the GACARAFE module, GP: the GACAP module. Results are reported on COCO \emph{val2017}.}}
	\resizebox{1.0\columnwidth}{!}{
		\begin{tabular}{cccccccccc}
			\toprule
			\textbf{MC} & \textbf{GE} & \textbf{GP} & $\mathrm{AP}^{bb}$ & AP & $\mathrm{AP}_{50}$ & $\mathrm{AP}_{75}$ & $\mathrm{AP}_{S}$ & $\mathrm{AP}_{M}$ & $\mathrm{AP}_{L}$ \\
			\midrule
			\multicolumn{3}{c}{FPN \cite{lin2017feature}} & 37.1 & 34.1 & 55.4 & 36.2 & 18.4 & 37.3 & 46.0 \\
			\multicolumn{3}{c}{PAFPN \cite{liu2018path}} & 37.6 & 34.4 & 55.9 & 36.4 & 18.7 & 37.5 & 47.2 \\
			\midrule
			$\checkmark$ &  &  & 38.6 & 35.4 & 57.4 & 37.5 & 19.5 & 38.6 & 48.3 \\
			& $\checkmark$ &  & 39.4 & 35.7 & 57.9 & 37.8 & 19.1  & 39.0 & 48.6 \\
			&  & $\checkmark$ & 38.3 & 35.0 & 56.6 & 37.2 & 18.0 & 38.4 & 48.3 \\
			\midrule
			$\checkmark$ & $\checkmark$ &  & 39.8 & 36.1 & \textbf{58.6} & \textbf{38.4} & \textbf{20.2} & \textbf{39.4} & 49.4 \\
			$\checkmark$ &  & $\checkmark$ & 38.9 & 35.5 & 57.7 & 37.6 & 19.0 & 39.2 & 48.8 \\
			& $\checkmark$ & $\checkmark$ & 39.6 & 35.9 & 58.2 & 37.9 & 19.8 & 39.0 & 48.9 \\
			\midrule
			$\checkmark$ & $\checkmark$ & $\checkmark$ & \textbf{40.0} & \textbf{36.2} & 58.4 & 38.1 & 20.1 & 39.2 & \textbf{49.6} \\
			\bottomrule
		\end{tabular}
	}
	\label{tab: effect of each module}
	\vspace{-10pt}
\end{table}

\subsection{Ablation Study}
\label{subsection: ablation study}
In this section, we conduct extensive ablation experiments to analyze the effects of main components in $A^2$-FPN. Note that the baseline method for all ablation studies is Mask R-CNN with R-50-PAFPN.

\textbf{Component-wise Analysis.} Firstly, we investigate the importance of each component in $A^2$-FPN. The baseline PAFPN is extended from FPN, where the difference from the original implementation in PANet \cite{liu2018path} is that we do not use Synchronized BatchNorm. And then MGC, GACARAFE, and GACAP are gradually applied to PAFPN by substituting the corresponding operations with them. 

As shown in Table \ref{tab: effect of each module}, each module effectively improves the performance of baseline. Specifically, the MGC module boosts both mask AP and box AP by 1.0\%. This benefits from that MGC sufficiently extracts discriminative features through cross-scale self-attention based on scaled cosine-similarity attention, and alleviates semantic information loss. The GACARAFE module contributes an improvement of 1.8\% and 1.3\% in terms of box AP and mask AP. Similarly, 0.7\% and 0.6\% for the GACAP module. These results indicate that content-aware sampling aggregates more semantic information and attention-guided fusion mitigates the inconsistency between adjacent levels. Besides, the GACARAFE module plays a key role in the construction of feature pyramid since there is only one top-down pathway but two bottom-up pathways.

When combining any two modules, the performance of baseline is further improved. For example, MGC and GACARAFE lead to a gain of 2.2\% box AP and 1.7\% mask AP. When integrating all three modules together, it achieves 40.0\% box AP and 36.2\% mask AP, with a gain of 2.4\% and 1.8\%. The improvements of $\mathrm{AP}_{S}$, $\mathrm{AP}_{M}$ and $\mathrm{AP}_{L}$ are 1.4\%, 1.7\% and 2.4\% respectively, suggesting that $A^2$-FPN is beneficial to various scales, especially the large scale.

\begin{table}[tp]
	\centering
	\caption{\small{\textbf{Ablation study of each module} on COCO \emph{val2017}. ``w/ ·'' indicates $A^2$-FPN only equipped with the module ·.}}
	\resizebox{1.0\columnwidth}{!}{
		\begin{tabular}{ccccccccc}
			\toprule
			Method & setting & $\mathrm{AP}^{bb}$ & AP & $\mathrm{AP}_{50}$ & $\mathrm{AP}_{75}$ & $\mathrm{AP}_{S}$ & $\mathrm{AP}_{M}$ & $\mathrm{AP}_{L}$ \\
			\midrule
			$A^2$-FPN & - & \textbf{40.0} & \textbf{36.2} & \textbf{58.4} & \textbf{38.1} & \textbf{20.1} & 39.2 & \textbf{49.6} \\
			$A^2$-FPN & w/o GCN & 39.6 & 35.9 & \textbf{58.4} & 38.0 & 19.2 & \textbf{39.4} & 49.0 \\
			\midrule
			w/ MGC & - & \textbf{38.6} & \textbf{35.4} & \textbf{57.4} & \textbf{37.5} & \textbf{19.5} & 38.6 & \textbf{48.3} \\
			w/ MGC & w/o GCN & 38.5 & 35.2 & 56.9 & \textbf{37.5} & 19.1 & \textbf{38.7} & 47.8 \\
			\midrule
			w/ GACARAFE & - & \textbf{39.4} & \textbf{35.7} & \textbf{57.9} & \textbf{37.8} & \textbf{19.1}  & \textbf{39.0} & 48.6 \\
			w/ CARAFE \cite{wang2019carafe} & - & 38.9 & 35.4 & 57.1 & 37.6 & 19.0 & 38.4 & \textbf{48.7} \\
			\midrule
			w/ GACAP & - & \textbf{38.3} & \textbf{35.0} & \textbf{56.6} & \textbf{37.2} & \textbf{18.0} & \textbf{38.4} & 48.3 \\
			w/ CAP & - & 38.0 & 34.8 & 56.2 & 37.0 & 17.7 & 38.0 & \textbf{48.5} \\
			\bottomrule
		\end{tabular}
	}
	\label{tab: ablation study of each module}
	\vspace{-10pt}
\end{table}

\def\visimgheighta{2.40cm}
\def\visimgheightb{2.39cm}

\begin{figure*}[tp]
	\centering
	\includegraphics[height=\visimgheighta]{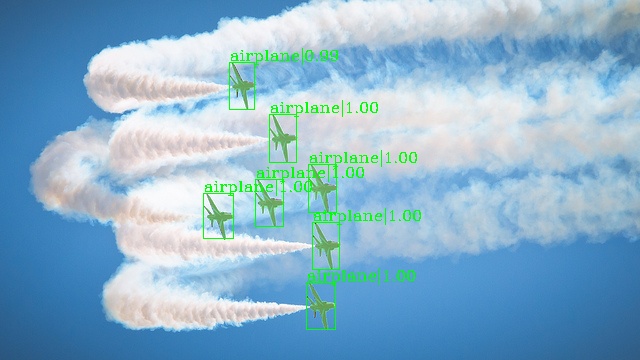}
	\includegraphics[height=\visimgheighta]{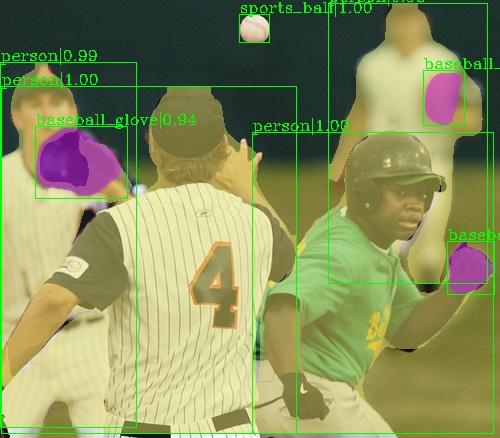}
	\includegraphics[height=\visimgheighta]{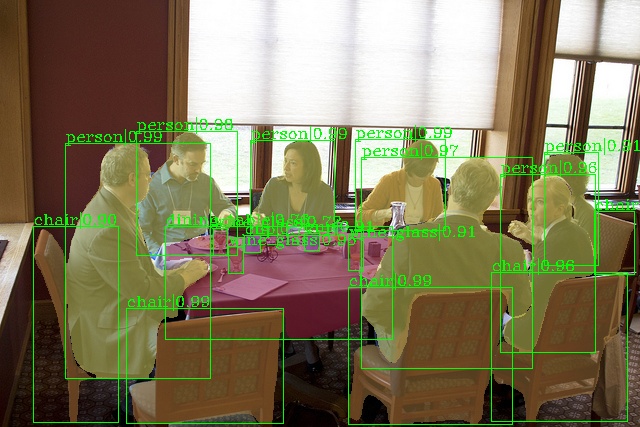}
	\includegraphics[height=\visimgheighta]{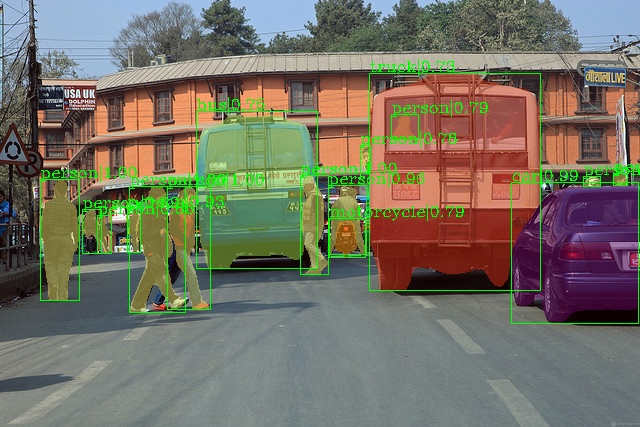}
	\includegraphics[height=\visimgheighta]{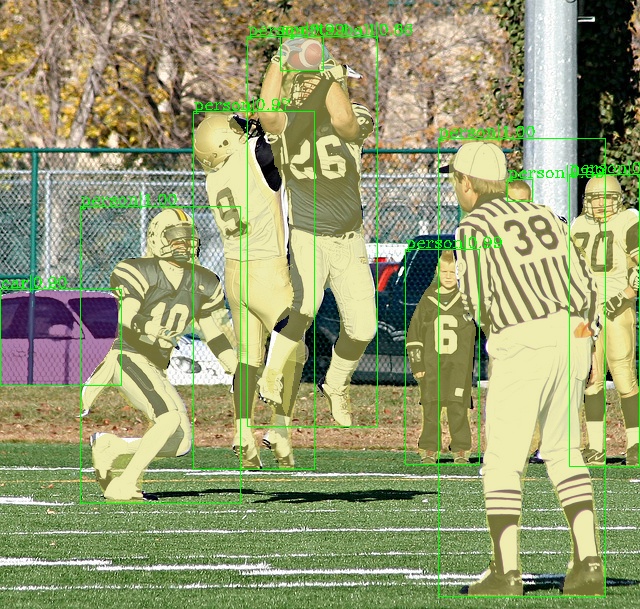}
	
	\includegraphics[height=\visimgheightb]{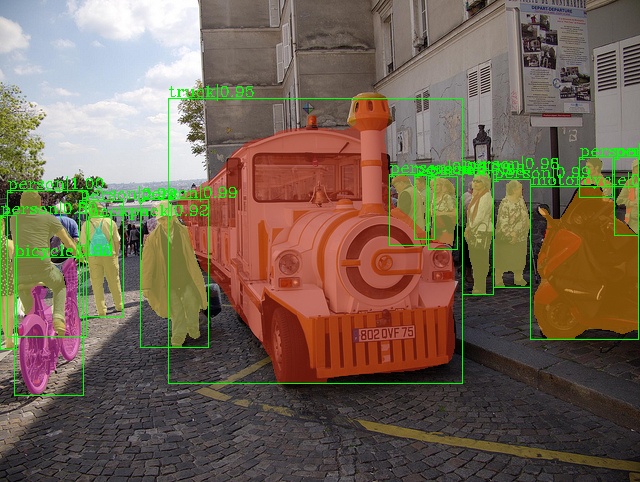}
	\includegraphics[height=\visimgheightb]{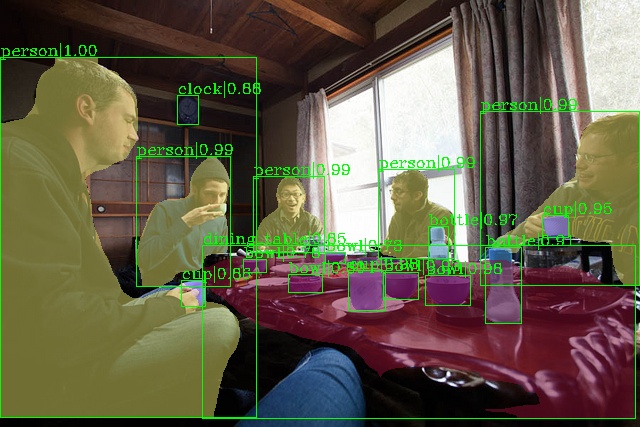}
	\includegraphics[height=\visimgheightb]{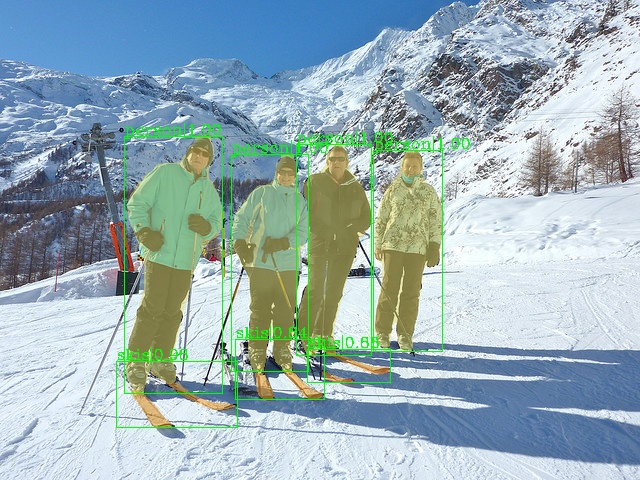}
	\includegraphics[height=\visimgheightb]{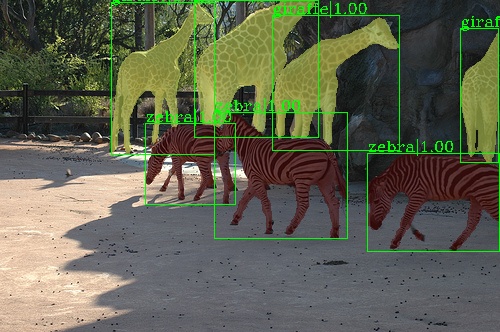}
	\includegraphics[height=\visimgheightb]{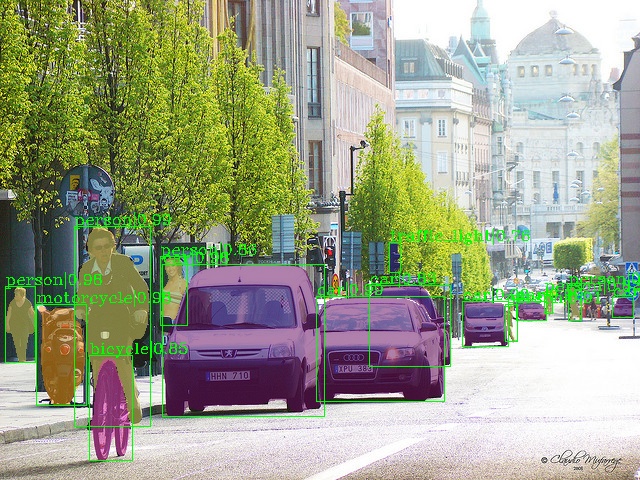}
	\caption{\small{Instance segmentation results of $\boldsymbol{A^2}$\textbf{-FPN} equipped with HTC on COCO \emph{val2017}.}}
	\label{fig: visualization}
	\vspace{-10pt}
\end{figure*}

\textbf{Ablation study of each module.} As shown in Table \ref{tab: ablation study of each module}, the performance of MGC and $A^2$-FPN degrades by 0.2\% and 0.3\% respectively when removing the GCNs. These results indicate that GCN can model the contextual relations between context features and benefit the subsequent feature fusion. Compared with CARAFE \cite{wang2019carafe}, GACARAFE further leads to a gain of 0.5\% box AP and 0.3\% mask AP. Similarly, 0.3\% box AP and 0.2\% mask AP for GACAP. These improvements benefit from that GACARAFE and GACAP aggregate complementary information from adjacent features for content-aware sampling and employ channel attention to enhance the semantic consistency.

\begin{table}[tp]
	\centering
	\caption{\small{\textbf{Impact of the number of context features} on COCO \emph{val2017}. $n_i$ means the number of context features collected from the i-th feature level $\mathcal{F}^{bb}_{i}$.}}
	\resizebox{1.0\columnwidth}{!}{
		\begin{tabular}{ccccccccc}
			\toprule
			Method & $n_i$ & $\mathrm{AP}^{bb}$ & AP & $\mathrm{AP}_{50}$ & $\mathrm{AP}_{75}$ & $\mathrm{AP}_{S}$ & $\mathrm{AP}_{M}$ & $\mathrm{AP}_{L}$ \\
			\midrule
			FPN \cite{lin2017feature} & - & 37.1 & 34.1 & 55.4 & 36.2 & 18.4 & 37.3 & 46.0 \\
			PAFPN \cite{liu2018path} & - & 37.6 & 34.4 & 55.9 & 36.4 & 18.7 & 37.5 & 47.2 \\
			\midrule
			w/ MGC & $ 64(i-1) $ & 38.7 & \textbf{35.4} & \textbf{57.8} & 37.4 & 19.2 & \textbf{38.8} & 48.1 \\
			w/ MGC & $ 160 $ & \textbf{38.8} & 35.3 & 57.4 & \textbf{37.6} & 19.2 & 38.7 & 48.2 \\
			w/ MGC & $ 64(6-i) $ & 38.6 & \textbf{35.4} & 57.4 & 37.5 & \textbf{19.5} & 38.6 & \textbf{48.3} \\
			\bottomrule
		\end{tabular}
	}
	\label{tab: the number of context features}
	\vspace{-5pt}
\end{table}

\textbf{Impact of the number of context features.} To investigate the impact of $n_i$, the number of context features collected from each level, we conduct contrast experiments with different settings. As shown in Table \ref{tab: the number of context features}, the MGC module is robust to the setting of number $n_i$ and achieves similar performance where the difference is only 0.1\% at most. Considering that the more context features from the high levels, the more parameters are introduced because of high dimension, we set the hyper-parameters as $ n_i = 64(6-i) $, i.e., $ n_i = \{256, 192, 128, 64\} $ and $ n = 640 $.

\begin{table}[tp]
	\centering
	\caption{\small{\textbf{Effectiveness of activation function $\boldsymbol{2 \cdot \mathrm{sigmoid}}$} on COCO \emph{val2017}. ``Act.'' is short for activation function, and $\sigma $ means the sigmoid function.}}
	\resizebox{1.0\columnwidth}{!}{
		\begin{tabular}{ccccccccc}
			\toprule
			Method & Act. & $\mathrm{AP}^{bb}$ & AP & $\mathrm{AP}_{50}$ & $\mathrm{AP}_{75}$ & $\mathrm{AP}_{S}$ & $\mathrm{AP}_{M}$ & $\mathrm{AP}_{L}$ \\
			\midrule
			FPN \cite{lin2017feature} & - & 37.1 & 34.1 & 55.4 & 36.2 & 18.4 & 37.3 & 46.0 \\
			PAFPN \cite{liu2018path} & - & 37.6 & 34.4 & 55.9 & 36.4 & 18.7 & 37.5 & 47.2 \\
			\midrule
			w/ GACARAFE & $ \sigma $ & 38.9 & 35.5 & 57.5 & 37.7 & \textbf{19.2} & 38.8 & 48.3 \\
			w/ GACARAFE & $ 2 \sigma$ & \textbf{39.4} & \textbf{35.7} & \textbf{57.9} & \textbf{37.8} & 19.1  & \textbf{39.0} & \textbf{48.6} \\
			\midrule
			w/ GACAP & $ \sigma $ & 38.2 & 34.8 & 56.3 & 37.0 & \textbf{18.3} & 38.3 & 47.7 \\
			w/ GACAP & $ 2 \sigma$ & \textbf{38.3} & \textbf{35.0} & \textbf{56.6} & \textbf{37.2} & 18.0 & \textbf{38.4} & \textbf{48.3} \\
			\bottomrule
		\end{tabular}
	}
	\label{tab: activation function}
	\vspace{-10pt}
\end{table}

\textbf{Effectiveness of activation function 2 $ \boldsymbol{\cdot}$ sigmoid.} To verify the effectiveness of activation function $2 \cdot \mathrm{sigmoid}$, we compare it with the conventional function $\mathrm{sigmoid}$ in both GACARAFE and GACAP. As shown in Table \ref{tab: activation function}, the GACARAFE module with function $2 \cdot \mathrm{sigmoid}$ achieves 39.4\% box AP and 35.7\% mask AP respectively, which is 0.5\% and 0.2\% higher than the counterpart with function $\mathrm{sigmoid}$. Meanwhile, function $2 \cdot \mathrm{sigmoid}$ contributes a gain of 0.1\% box AP and 0.2\% mask AP in the GACAP module. The pyramidal features are iteratively reweighted in the feature fusion, and the final calibration weights are the product of several activated weights. When using $\mathrm{sigmoid}$ as the activation function, the mean of channel-wise weights would be always less than 1 and decay exponentially after successively reweighting, so that pyramidal features would be suppressed in the iterative fusion. On the contrary, the activation function $2 \cdot \mathrm{sigmoid}$ can keep the mean of channel-wise weights being 1 and excite or restrain features selectively.

\begin{table}[tp]
	\centering
	\caption{\small{\textbf{Complexity analysis of $\boldsymbol{A^2}$-FPN} on COCO \emph{val2017}.}}
	\resizebox{1.0\columnwidth}{!}{
		\begin{tabular}{ccccccc}
			\toprule
			Method & Image Size & \#FLOPs & \#Params & FPS & $\mathrm{AP}^{bb}$ & AP \\
			\midrule
			FPN \cite{lin2017feature} & $1280 \times 832$ & \textbf{283.28G} & \textbf{44.18M} & \textbf{13.1} & 37.1 & 34.1 \\
			PAFPN \cite{liu2018path} & $1280 \times 832$ & 309.05G & 47.72M & 12.8 & 37.6 & 34.4 \\
			\midrule
			$A^2$-FPN & $1280 \times 832$ & 375.39G & 57.49M & 9.5 & \textbf{40.0} & \textbf{36.2} \\
			$A^2$-FPN-Lite & $1280 \times 832$ & 319.91G & 48.83M & 11.1 & 39.6 & 36.0 \\
			\bottomrule
		\end{tabular}
	}
	\label{tab: complexity analysis}
	\vspace{-10pt}
\end{table}

\subsection{Complexity Analysis}
\label{subsection: complexity analysis}
As shown in Table \ref{tab: complexity analysis}, we analyze the complexity of $A^2$-FPN. $A^2$-FPN adds some computation cost, but $A^2$-FPN-Lite can reduce it greatly at a small performance sacrifice. Compared to PAFPN, $A^2$-FPN-Lite improves the performance significantly while increases the computation complexity acceptably, proving that the improvement mostly comes from our attention-guided feature aggregation.

\section{Conclusion}
\label{section: conclusion}
We propose Attention Aggregation based Feature Pyramid Network ($A^2$-FPN), which improves multi-scale feature learning through attention-guided feature aggregation. $A^2$-FPN extracts discriminative features by collecting-distributing multi-level global context features, and fuses adjacent levels using content-aware sampling and channel-wise reweighting before element-wise addition. Upon various backbone networks and instance segmentation frameworks, $A^2$-FPN consistently and substantially improves the performance of baseline methods.

%\section*{Acknowledgments}
\vspace{-12pt}
\paragraph{Acknowledgments.}
This work was supported by the National Natural Science Foundation of China under Grant No. 61771288, Cross-Media Intelligent Technology Project of Beijing National Research Center for Information Science and Technology (BNRist) under Grant No. BNR2019TD01022 and the research fund under Grant No. 2019GQG0001 from the Institute for Guo Qiang, Tsinghua University. We also thank for the funding support of Huawei Technologies Co. Ltd.

{\small
\bibliographystyle{ieee_fullname}
\bibliography{egbib}

\begin{thebibliography}{10}\itemsep=-1pt

\bibitem{ba2016layer}
Jimmy~Lei Ba, Jamie~Ryan Kiros, and Geoffrey~E Hinton.
\newblock Layer normalization.
\newblock {\em arXiv preprint arXiv:1607.06450}, 2016.

\bibitem{bai2017deep}
Min Bai and Raquel Urtasun.
\newblock Deep watershed transform for instance segmentation.
\newblock In {\em Proceedings of the IEEE Conference on Computer Vision and
  Pattern Recognition}, pages 5221--5229, 2017.

\bibitem{cai2018cascade}
Zhaowei Cai and Nuno Vasconcelos.
\newblock Cascade r-cnn: Delving into high quality object detection.
\newblock In {\em Proceedings of the IEEE conference on computer vision and
  pattern recognition}, pages 6154--6162, 2018.

\bibitem{cao2019gcnet}
Yue Cao, Jiarui Xu, Stephen Lin, Fangyun Wei, and Han Hu.
\newblock Gcnet: Non-local networks meet squeeze-excitation networks and
  beyond.
\newblock In {\em Proceedings of the IEEE International Conference on Computer
  Vision Workshops}, pages 0--0, 2019.

\bibitem{chen2019hybrid}
Kai Chen, Jiangmiao Pang, Jiaqi Wang, Yu Xiong, Xiaoxiao Li, Shuyang Sun,
  Wansen Feng, Ziwei Liu, Jianping Shi, Wanli Ouyang, et~al.
\newblock Hybrid task cascade for instance segmentation.
\newblock In {\em Proceedings of the IEEE conference on computer vision and
  pattern recognition}, pages 4974--4983, 2019.

\bibitem{mmdetection}
Kai Chen, Jiaqi Wang, Jiangmiao Pang, Yuhang Cao, Yu Xiong, Xiaoxiao Li,
  Shuyang Sun, Wansen Feng, Ziwei Liu, Jiarui Xu, Zheng Zhang, Dazhi Cheng,
  Chenchen Zhu, Tianheng Cheng, Qijie Zhao, Buyu Li, Xin Lu, Rui Zhu, Yue Wu,
  Jifeng Dai, Jingdong Wang, Jianping Shi, Wanli Ouyang, Chen~Change Loy, and
  Dahua Lin.
\newblock {MMDetection}: Open mmlab detection toolbox and benchmark.
\newblock {\em arXiv preprint arXiv:1906.07155}, 2019.

\bibitem{chen2019graph}
Yunpeng Chen, Marcus Rohrbach, Zhicheng Yan, Yan Shuicheng, Jiashi Feng, and
  Yannis Kalantidis.
\newblock Graph-based global reasoning networks.
\newblock In {\em Proceedings of the IEEE Conference on Computer Vision and
  Pattern Recognition}, pages 433--442, 2019.

\bibitem{dai2016instance}
Jifeng Dai, Kaiming He, and Jian Sun.
\newblock Instance-aware semantic segmentation via multi-task network cascades.
\newblock In {\em Proceedings of the IEEE Conference on Computer Vision and
  Pattern Recognition}, pages 3150--3158, 2016.

\bibitem{de2017semantic}
Bert De~Brabandere, Davy Neven, and Luc Van~Gool.
\newblock Semantic instance segmentation with a discriminative loss function.
\newblock {\em arXiv preprint arXiv:1708.02551}, 2017.

\bibitem{fathi2017semantic}
Alireza Fathi, Zbigniew Wojna, Vivek Rathod, Peng Wang, Hyun~Oh Song, Sergio
  Guadarrama, and Kevin~P Murphy.
\newblock Semantic instance segmentation via deep metric learning.
\newblock {\em arXiv preprint arXiv:1703.10277}, 2017.

\bibitem{fu2019dual}
Jun Fu, Jing Liu, Haijie Tian, Yong Li, Yongjun Bao, Zhiwei Fang, and Hanqing
  Lu.
\newblock Dual attention network for scene segmentation.
\newblock In {\em Proceedings of the IEEE Conference on Computer Vision and
  Pattern Recognition}, pages 3146--3154, 2019.

\bibitem{ghiasi2019fpn}
Golnaz Ghiasi, Tsung-Yi Lin, and Quoc~V Le.
\newblock Nas-fpn: Learning scalable feature pyramid architecture for object
  detection.
\newblock In {\em Proceedings of the IEEE conference on computer vision and
  pattern recognition}, pages 7036--7045, 2019.

\bibitem{guo2020augfpn}
Chaoxu Guo, Bin Fan, Qian Zhang, Shiming Xiang, and Chunhong Pan.
\newblock Augfpn: Improving multi-scale feature learning for object detection.
\newblock In {\em Proceedings of the IEEE/CVF Conference on Computer Vision and
  Pattern Recognition}, pages 12595--12604, 2020.

\bibitem{he2017mask}
Kaiming He, Georgia Gkioxari, Piotr Doll{\'a}r, and Ross Girshick.
\newblock Mask r-cnn.
\newblock In {\em Proceedings of the IEEE international conference on computer
  vision}, pages 2961--2969, 2017.

\bibitem{he2016deep}
Kaiming He, Xiangyu Zhang, Shaoqing Ren, and Jian Sun.
\newblock Deep residual learning for image recognition.
\newblock In {\em Proceedings of the IEEE conference on computer vision and
  pattern recognition}, pages 770--778, 2016.

\bibitem{hu2018squeeze}
Jie Hu, Li Shen, and Gang Sun.
\newblock Squeeze-and-excitation networks.
\newblock In {\em Proceedings of the IEEE conference on computer vision and
  pattern recognition}, pages 7132--7141, 2018.

\bibitem{huang2019mask}
Zhaojin Huang, Lichao Huang, Yongchao Gong, Chang Huang, and Xinggang Wang.
\newblock Mask scoring r-cnn.
\newblock In {\em Proceedings of the IEEE conference on computer vision and
  pattern recognition}, pages 6409--6418, 2019.

\bibitem{kirillov2017instancecut}
Alexander Kirillov, Evgeny Levinkov, Bjoern Andres, Bogdan Savchynskyy, and
  Carsten Rother.
\newblock Instancecut: from edges to instances with multicut.
\newblock In {\em Proceedings of the IEEE Conference on Computer Vision and
  Pattern Recognition}, pages 5008--5017, 2017.

\bibitem{li2018beyond}
Yin Li and Abhinav Gupta.
\newblock Beyond grids: Learning graph representations for visual recognition.
\newblock In {\em Advances in Neural Information Processing Systems}, pages
  9225--9235, 2018.

\bibitem{li2017fully}
Yi Li, Haozhi Qi, Jifeng Dai, Xiangyang Ji, and Yichen Wei.
\newblock Fully convolutional instance-aware semantic segmentation.
\newblock In {\em Proceedings of the IEEE Conference on Computer Vision and
  Pattern Recognition}, pages 2359--2367, 2017.

\bibitem{liang2018symbolic}
Xiaodan Liang, Zhiting Hu, Hao Zhang, Liang Lin, and Eric~P Xing.
\newblock Symbolic graph reasoning meets convolutions.
\newblock In {\em Advances in Neural Information Processing Systems}, pages
  1853--1863, 2018.

\bibitem{lin2017feature}
Tsung-Yi Lin, Piotr Doll{\'a}r, Ross Girshick, Kaiming He, Bharath Hariharan,
  and Serge Belongie.
\newblock Feature pyramid networks for object detection.
\newblock In {\em Proceedings of the IEEE conference on computer vision and
  pattern recognition}, pages 2117--2125, 2017.

\bibitem{lin2017focal}
Tsung-Yi Lin, Priya Goyal, Ross Girshick, Kaiming He, and Piotr Doll{\'a}r.
\newblock Focal loss for dense object detection.
\newblock In {\em Proceedings of the IEEE international conference on computer
  vision}, pages 2980--2988, 2017.

\bibitem{lin2014microsoft}
Tsung-Yi Lin, Michael Maire, Serge Belongie, James Hays, Pietro Perona, Deva
  Ramanan, Piotr Doll{\'a}r, and C~Lawrence Zitnick.
\newblock Microsoft coco: Common objects in context.
\newblock In {\em European conference on computer vision}, pages 740--755.
  Springer, 2014.

\bibitem{liu2017sgn}
Shu Liu, Jiaya Jia, Sanja Fidler, and Raquel Urtasun.
\newblock Sgn: Sequential grouping networks for instance segmentation.
\newblock In {\em Proceedings of the IEEE International Conference on Computer
  Vision}, pages 3496--3504, 2017.

\bibitem{liu2018path}
Shu Liu, Lu Qi, Haifang Qin, Jianping Shi, and Jiaya Jia.
\newblock Path aggregation network for instance segmentation.
\newblock In {\em Proceedings of the IEEE conference on computer vision and
  pattern recognition}, pages 8759--8768, 2018.

\bibitem{liu2016ssd}
Wei Liu, Dragomir Anguelov, Dumitru Erhan, Christian Szegedy, Scott Reed,
  Cheng-Yang Fu, and Alexander~C Berg.
\newblock Ssd: Single shot multibox detector.
\newblock In {\em European conference on computer vision}, pages 21--37.
  Springer, 2016.

\bibitem{liu2015parsenet}
Wei Liu, Andrew Rabinovich, and Alexander~C Berg.
\newblock Parsenet: Looking wider to see better.
\newblock {\em arXiv preprint arXiv:1506.04579}, 2015.

\bibitem{newell2017associative}
Alejandro Newell, Zhiao Huang, and Jia Deng.
\newblock Associative embedding: End-to-end learning for joint detection and
  grouping.
\newblock In {\em Advances in neural information processing systems}, pages
  2277--2287, 2017.

\bibitem{pinheiro2015learning}
Pedro~OO Pinheiro, Ronan Collobert, and Piotr Doll{\'a}r.
\newblock Learning to segment object candidates.
\newblock In {\em Advances in Neural Information Processing Systems}, pages
  1990--1998, 2015.

\bibitem{pinheiro2016learning}
Pedro~O Pinheiro, Tsung-Yi Lin, Ronan Collobert, and Piotr Doll{\'a}r.
\newblock Learning to refine object segments.
\newblock In {\em European conference on computer vision}, pages 75--91.
  Springer, 2016.

\bibitem{redmon2018yolov3}
Joseph Redmon and Ali Farhadi.
\newblock Yolov3: An incremental improvement.
\newblock {\em arXiv preprint arXiv:1804.02767}, 2018.

\bibitem{ren2015faster}
Shaoqing Ren, Kaiming He, Ross Girshick, and Jian Sun.
\newblock Faster r-cnn: Towards real-time object detection with region proposal
  networks.
\newblock In {\em Advances in neural information processing systems}, pages
  91--99, 2015.

\bibitem{shi2016real}
Wenzhe Shi, Jose Caballero, Ferenc Husz{\'a}r, Johannes Totz, Andrew~P Aitken,
  Rob Bishop, Daniel Rueckert, and Zehan Wang.
\newblock Real-time single image and video super-resolution using an efficient
  sub-pixel convolutional neural network.
\newblock In {\em Proceedings of the IEEE conference on computer vision and
  pattern recognition}, pages 1874--1883, 2016.

\bibitem{tan2020efficientdet}
Mingxing Tan, Ruoming Pang, and Quoc~V Le.
\newblock Efficientdet: Scalable and efficient object detection.
\newblock In {\em Proceedings of the IEEE/CVF Conference on Computer Vision and
  Pattern Recognition}, pages 10781--10790, 2020.

\bibitem{vaswani2017attention}
Ashish Vaswani, Noam Shazeer, Niki Parmar, Jakob Uszkoreit, Llion Jones,
  Aidan~N Gomez, {\L}ukasz Kaiser, and Illia Polosukhin.
\newblock Attention is all you need.
\newblock In {\em Advances in neural information processing systems}, pages
  5998--6008, 2017.

\bibitem{wang2019carafe}
Jiaqi Wang, Kai Chen, Rui Xu, Ziwei Liu, Chen~Change Loy, and Dahua Lin.
\newblock Carafe: Content-aware reassembly of features.
\newblock In {\em Proceedings of the IEEE International Conference on Computer
  Vision}, pages 3007--3016, 2019.

\bibitem{wang2018non}
Xiaolong Wang, Ross Girshick, Abhinav Gupta, and Kaiming He.
\newblock Non-local neural networks.
\newblock In {\em Proceedings of the IEEE conference on computer vision and
  pattern recognition}, pages 7794--7803, 2018.

\bibitem{wang2020attentive}
Yi Wang, Ying-Cong Chen, Xiangyu Zhang, Jian Sun, and Jiaya Jia.
\newblock Attentive normalization for conditional image generation.
\newblock In {\em Proceedings of the IEEE/CVF Conference on Computer Vision and
  Pattern Recognition}, pages 5094--5103, 2020.

\bibitem{xie2017aggregated}
Saining Xie, Ross Girshick, Piotr Doll{\'a}r, Zhuowen Tu, and Kaiming He.
\newblock Aggregated residual transformations for deep neural networks.
\newblock In {\em Proceedings of the IEEE conference on computer vision and
  pattern recognition}, pages 1492--1500, 2017.

\bibitem{zagoruyko2016multipath}
Sergey Zagoruyko, Adam Lerer, Tsung-Yi Lin, Pedro~O Pinheiro, Sam Gross,
  Soumith Chintala, and Piotr Doll{\'a}r.
\newblock A multipath network for object detection.
\newblock {\em arXiv preprint arXiv:1604.02135}, 2016.

\bibitem{zhang2019latentgnn}
Songyang Zhang, Xuming He, and Shipeng Yan.
\newblock Latentgnn: Learning efficient non-local relations for visual
  recognition.
\newblock In {\em International Conference on Machine Learning}, pages
  7374--7383, 2019.

\end{thebibliography}
}

\end{document}